\documentclass[conference]{IEEEtran}
\IEEEoverridecommandlockouts
\usepackage{cite}
\usepackage{amsmath,amssymb,amsfonts}
\usepackage{algorithmic}
\usepackage{graphicx}
\usepackage{textcomp}
\usepackage{xcolor}
\def\BibTeX{{\rm B\kern-.05em{\sc i\kern-.025em b}\kern-.08em
    T\kern-.1667em\lower.7ex\hbox{E}\kern-.125emX}}
\begin{document}

\title{Molecular Odor Prediction Based on Multi-Feature Graph Attention Networks}

\author{
    HongXin Xie$^1$, JianDe Sun$^1$, Yi Shao$^1$, Shuai Li$^2$, Sujuan Hou$^1$, YuLong Sun$^1$, Jian Wang$^1$ \\
$^1$ Shandong Normal University, Jinan, China \\
$^2$ Shandong University, Jinan, China \\
jiandesun@hotmail.com
}

\maketitle

\begin{abstract}
Olfactory perception plays a critical role in both human and organismal interactions, yet understanding of its underlying mechanisms and influencing factors remain insufficient. Molecular structures influence odor perception through intricate biochemical interactions, and accurately quantifying structure-odor relationships presents significant challenges. The Quantitative Structure-Odor Relationship (QSOR) task, which involves predicting the associations between molecular structures and their corresponding odors, seeks to address these challenges. To this end, we propose a method for QSOR, utilizing Graph Attention Networks to model molecular structures and capture both local and global features. Unlike conventional QSOR approaches reliant on predefined descriptors, our method leverages diverse molecular feature extraction techniques to automatically learn comprehensive representations. This integration enhances the model's capacity to handle complex molecular information, improves prediction accuracy. Our approach demonstrates clear advantages in QSOR prediction tasks, offering valuable insights into the application of deep learning in cheminformatics.
\end{abstract}


\section{INTRODUCTION}
\label{sec:intro}

Olfaction, an extraordinary form of chemical perception, plays a vital role in human sensory experiences, complementing vision and hearing. Despite advances in understanding its physiological basis, the quantitative mapping of molecular structures to olfactory perception remains a significant challenge\cite{2}. The Quantitative Structure-Odor Relationship (QSOR) aims to uncover how molecular structures encode olfactory characteristics. It bridges chemistry, neuroscience, and sensory science, providing insights into sensory mechanisms and applications in fields such as food, fragrance, and materials science. Olfactory perception is inherently complex, influenced by multidimensional and multi-scale factors ranging from atomic-level bond properties to three-dimensional molecular structures. Traditional QSOR methods\cite{3,x1,x2}, limited by scarce experimental data and simplistic statistical models, struggle to capture the diversity and intricacies of odor perception\cite{2}. To address this, we propose a graph neural network framework that integrates multi-scale molecular features, including atomic-level attributes and global topological properties. Our approach significantly outperforms existing models in predicting odor descriptors, as demonstrated by experimental results. QSOR datasets often exhibit sparse and imbalanced odor labels, where rare odor categories are underrepresented. To mitigate this, we developed a dynamic loss function that adaptively prioritizes rare odors during training, effectively addressing label imbalance and improving predictive performance, as validated through extensive experiments. While molecular structure remains central to QSOR, odor perception is further shaped by spatial configurations and electronic interactions. Existing QSOR models, which rely on static molecular data, fail to capture these dynamic interactions\cite{x3}. To overcome this, we designed a method combining multi-level feature integration and attentional aggregation mechanisms. This design enables accurate modeling of molecular dynamics, significantly enhancing the predictive accuracy of odor-structure relationships. Specifically, the main contributions of our work are as follows:

\begin{itemize}
  \item We propose a multi-level feature extraction method to address the limitations of existing approaches in capturing the intricate relationships between molecular structures and odors, with its effectiveness validated by experimental results.
  \item We propose an attention-based node embedding aggregation mechanism that overcomes the limitations of traditional pooling methods. Compared to existing approaches, it introduces a dynamic prioritization strategy for relevant molecular features.
  \item We propose a network architecture based on graph attention convolution, which alleviates the limitations of traditional GNNs in modeling complex structural and relational features, thereby improving the accuracy of molecular odor prediction.
  \item To tackle label imbalance in multi-label classification, we introduce the Adaptive Focal Loss function, which dynamically adjusts training weights to focus on hard-to-classify samples, significantly improving rare label predictions.
\end{itemize}

\begin{figure*}[htbp]
\centerline{\includegraphics[width=1\textwidth,height=0.5\textheight,keepaspectratio]{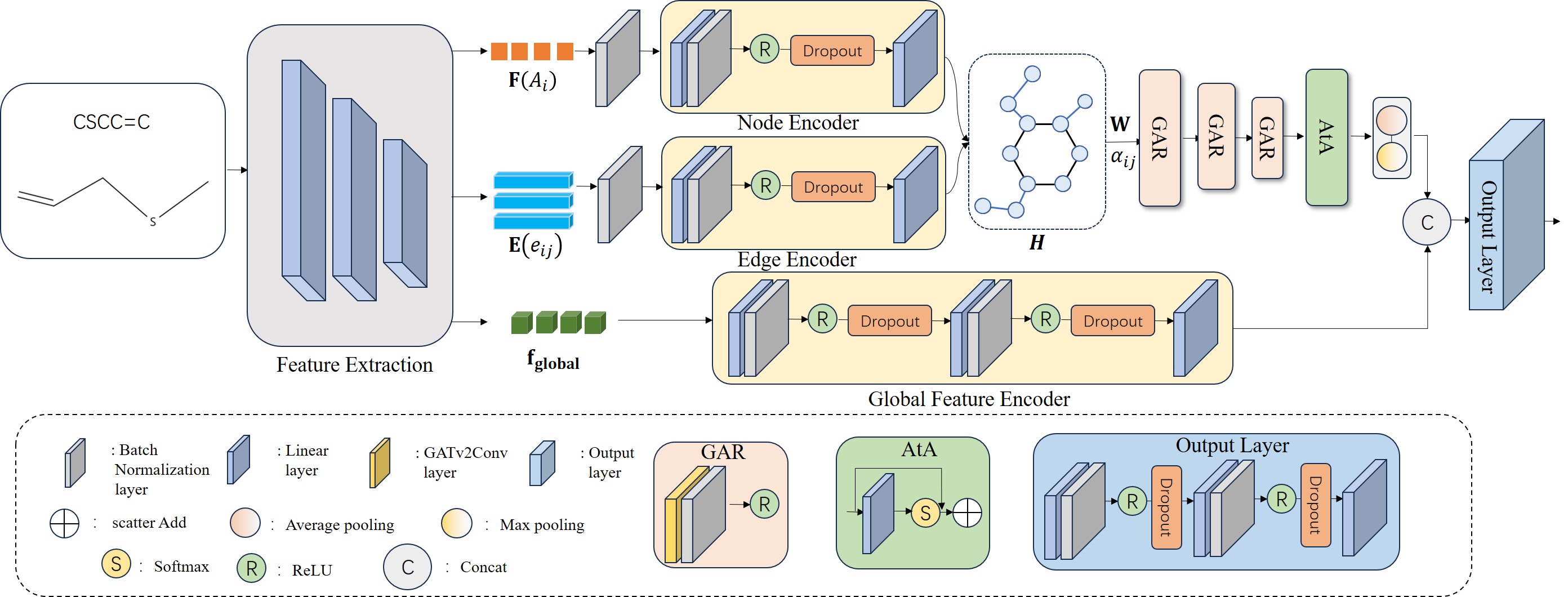}}
\caption{Overview of our proposed architecture.Our network architecture demonstrates a multi-layer feature extraction and graph attention aggregation framework for efficiently encoding and learning representations of complex input data.}
\label{fig}
\end{figure*}

\section{PROPOSED METHOD}

To address the complexities of molecular odor prediction, as shown in Fig. 1, we proposes a predictive method based on a multi-layer graph attention network. By integrating multi-level molecular feature extraction with  graph attentional aggregation modeling and designing a dynamic loss function, the method effectively captures the complex relationships between molecular structures and odor attributes.

\subsection{Multi-level Feature Extraction}

We propose a multi-level feature extraction approach that employs multidimensional representations of molecular structures to effectively capture the latent relationships between molecules and odors.

\noindent\textbf{Local Characteristics} To accurately capture molecular characteristics, we developed a comprehensive local feature extraction framework that integrates atomic-level and bond-level information to represent the intricate structural and chemical interactions influencing odor profiles. At the atomic level, fundamental descriptors and advanced chemical properties are utilized, with systematic normalization applied to ensure consistent scaling and numerical stability, resulting in the atomic features \(\mathbf{F}(A_i)\). Fundamental descriptors, such as atomic number, degree, formal charge, and radical count, provide baseline information on atomic composition and connectivity. Electronegativity reflects an atom’s ability to attract electrons, influencing molecular polarity and receptor binding, which are critical for odor detection in polar functional groups like hydroxyl\cite{5}. Atomic volume captures spatial and conformational characteristics of atoms, crucial for determining molecular volatility and diffusivity, both essential for odor perception\cite{6}. Electron affinity quantifies an atom’s capacity to accept electrons, offering insights into molecular reactivity and stability, directly impacting odor persistence and chemical transformation\cite{7}. Advanced properties such as electronegativity, atomic volume, and electron affinity are normalized using a unified formula:

\begin{equation}
\mathbf{f}_{X(A_{i})}=\frac{X(A_{i})-\mathbf{X}_{\min}}{\mathbf{X}_{\max}-\mathbf{X}_{\min}}
\end{equation}
where \(X =\)\(\{\)\(electronegativity,volume,electron affinity\)\(\}\), \(X(A_{i})\) is its value for atom \(A_{i}\), and \(\mathbf{X}_{\min}\) and \(\mathbf{X}_{\max}\) denote the normalization bounds. For these properties, the bounds are defined as follows: electronegativity (0.8, 4.0), atomic volume (4.0, 46.0), and electron affinity (-70.0, 350.0).

In addition to atomic-level features, bond-level characteristics \(\mathbf{E}(e_{ij})\) are incorporated to comprehensively represent molecular connectivity. These include bond type, conjugation, aromaticity, and branching, as well as the proportions of single, double, and triple bonds. Such features influence electronic distribution and molecular geometry, which are critical in determining key odor-related properties, such as volatility, polarity, and receptor binding modes\cite{8}. The bond feature vector \(\mathbf{f}(e_{ij})\) is defined as:

\begin{equation}
\mathbf{f}(e_{ij})=[\mathbf{b},\delta,\phi]
\end{equation}
where \(\mathbf{b}\) represents the basic properties of the edge, including key type etc. \(\delta=[\delta_i,\delta_j]\) represent the branching degree of atoms at both ends of the bond, that is, the number of adjacent atoms connected by each atom, and \(\phi=[\phi_1,\phi_2,\phi_3]\) represents the ratio of single bond, double bond and triple bond in a molecule.

Functional group features, extracted using a SMARTS\cite{x9}-based recognition approach, add another layer of detail by identifying chemical substructures that directly impact molecular polarity and hydrophobicity. These features, crucial for receptor interactions and volatility, enable the model to recognize patterns linked to specific odor characteristics\cite{9}. Given a molecule \(\mathcal{M}\) and a set of SMARTS patterns \(G=[G_1,G_2,…,G_n]\), the feature vector is:
\begin{equation}
f_i=\begin{cases}1,&\text{if }\mathcal{M}\text{ matches }G_i\\0,&\text{ otherwise }\end{cases}
\end{equation}
Thus, \(f_i\) is a Boolean value indicating the presence of functional group \(G_i\) in molecule \(\mathcal{M}\), determined by SMARTS pattern matching.

The functional group feature vector is defined as:
\begin{equation}
\mathbf{f}_{\mathrm{func}}(A_i)=[f_1,f_2,...,f_n]
\end{equation}
where \(\mathbf{f}_{\mathrm{func}}(A_i)\) denotes the functional group feature. 

Finally, molecules are modeled as graphs with atoms as nodes and bonds as edges. By representing molecular structures as graphs, the constructed feature representations enable the graph neural network to effectively model the nodes (atoms) and edges (chemical bonds) within the molecule, capturing its intricate relationships and complex characteristics.

\noindent\textbf{Global Characteristics} We comprehensively represented molecular structures by integrating multiple molecular fingerprints, including Morgan fingerprints\cite{10}, MACCS fingerprints\cite{11}, and Topological fingerprints\cite{12}. Each fingerprint captures distinct structural aspects. Morgan fingerprints are effective for encoding local topological features by hashing atomic environments and their neighbors into fixed-length bit vectors. MACCS keys focus on molecular scaffolds, consisting of 166 predefined substructure features, where every bit represents the presence or absence of a specific chemical fragment. This compact representation highlights critical structural patterns in chemistry. Topological fingerprints, on the other hand, capture global molecular topology by computing the shortest paths between all atomic pairs, effectively encoding the molecular distance matrix. The feature vectors for Morgan, MACCS, and Topological fingerprints are denoted as \(\mathbf{f_{Morgan}}\) \(\in\) \(R^{d_1}\), \(\mathbf{f_{MACCS}}\) \(\in\) \(R^{d_2}\), \(\mathbf{f_{Topo}}\) \(\in\) \(R^{d_3}\), respectively, where \(d_1,d_2,d_3\) are their respective dimensions. The global features vector \(\mathbf{f_{global}}\) is expressed as:
\begin{equation}
\mathbf{f_{global}}=\mathbf{f_{Morgan}}\parallel\mathbf{f_{MACCS}}\parallel\mathbf{f_{Topo}}
\end{equation}
where \(||\) denotes the concatenation of vectors along their dimensions, resulting in \(\mathbf{f}_{\mathbf{global}}\) \(\in\) \(R^{d_1+d_2+d_3}\)

By integrating these local and global characteristics, we achieved a multi-level representation of molecular structures, significantly enhancing the model's performance in odor prediction tasks.

\subsection{Graph Attentional Aggregation}

\noindent\textbf{Our Graph Network} We adopt a structure based on a three-layer Graph Attention Network (GAT) architecture\cite{4}, designed to progressively capture node and edge information in molecular graphs. The first layer initializes the model’s convolutional operations, receiving encoded node and edge features. Message passing is performed using a multi-head attention mechanism with \(K=4\) attention heads. The single-head attention-based convolution operation is defined as:
\begin{equation}
h_i^{(l+1)}=\sum_{j\in\mathcal{N}(i)}\alpha_{ij}^{(l)}W^{(l)}h_j^{(l)}
\end{equation}
where \(h_i^{(l)}\) is the feature of node \(i\) at layer \(l\), \(W^{(l)}\) is a learnable weight matrix, \(\alpha_{ij}^{(l)}\) is the attention coefficient between node \(i\) and its neighbor \(j\), and \(\mathcal{N}(i)\) is the set of neighbors of \(i\). The attention coefficient \(a_{ij}^{(l)}\) is computed as:
 
{\scriptsize
\begin{equation}
\alpha_{ij}^{(l)}=\frac{\exp\left(\text{LeakyReLU}\left(a^\top\left[W^{(l)}h_i^{(l)}\parallel W^{(l)}h_j^{(l)}\parallel e_{ij}\right]\right)\right)}{\sum_{k\in\mathcal{N}(i)}\exp\left(\text{LeakyReLU}\left(a^\top\left[W^{(l)}h_i^{(l)}\parallel W^{(l)}h_k^{(l)}\parallel e_{ik}\right]\right)\right)}
\end{equation}
}
where \(e_{ij}\) represents the edge feature between nodes \(i\) and \(j\), \(a^\top\) is a learnable attention vector, and \(||\) denotes vector concatenation. For \(k\) head attention, the node feature update becomes:
\begin{equation}
h_i^{(l+1)}=\|_{k=1}^K\sum_{j\in\mathcal{N}(i)}\alpha_{ij}^kW^kh_j^{(l)}
\end{equation}
where \(||\) denotes concatenation across attention heads.

The multi-head attention mechanism captures diverse neighborhood interactions by dynamically assigning weights based on node features, improving the model's expressiveness and stabilizing training with Batch Normalization and ReLU activation. The second layer builds on the first, capturing more complex atom-bond interactions and expanding the receptive field to integrate local and global information, with Batch Normalization and ReLU ensuring robust training. The final layer refines node features using a single attention head, consolidating information from previous layers without introducing new data, while retaining Batch Normalization and ReLU to preserve nonlinearity and smooth feature transitions. This hierarchical design progressively captures molecular graph information, balancing local and global structural representations, and effectively leveraging the interplay between atomic and bond-level features.

\noindent\textbf{Attention Integration} To effectively extract and integrate global information from molecular graphs, we introduce the Attention-based Aggregation (AtA) module in conjunction with multi-pooling operations. Unlike traditional graph convolutional networks\cite{graph} that rely on simple mean or max pooling, our approach utilizes attention mechanisms to dynamically assign importance weights to nodes, enabling the model to prioritize critical local structures and capture deeper dependencies. This enhances the expressiveness of global molecular representations. The attention weight \(\alpha_i\) for node \(i\) is calculated as:
\begin{equation}
\alpha_i=\frac{\exp(W_\mathrm{att}\cdot x_i)}{\sum_{j\in N}\exp(W_\mathrm{att}\cdot x_j)}
\end{equation}
where \(W_\mathrm{att}\) is the learnable attention weight matrix, \(x_i\) is the feature vector of node \(i\), and \(N\) represents all nodes in the current batch. These weights allow the model to emphasize nodes most relevant to specific tasks, overcoming the limitations of traditional pooling methods that treat all nodes equally.
Global feature aggregation is achieved by summing node features weighted by their attention scores:
\begin{equation}
h'=\sum_i\alpha_ix_i
\end{equation}
where \(h'\) represents the aggregated global feature, dynamically capturing the contributions of critical nodes. For graph \(G_k\), the global feature representation is computed as:
\begin{equation}
H_k=\sum_{i\in G_k}\alpha_ix_i
\end{equation}
Here, \(H_k\) represents the global feature of the graph \(G_k\), dynamically emphasizing critical local structures.

To enhance global representation, we employ global average and max pooling, capturing both macro-level and critical local molecular features. These pooled results, along with global molecular characteristics integrated via a multilayer perceptron (MLP), are fused into a unified representation that balances local and global information. This approach combines attention aggregation, multi-pooling, and feature fusion to improve model robustness and accuracy, particularly in predicting molecular odor properties.

\subsection{Loss Function}

In multi-label molecular odor prediction tasks, data often exhibit significant imbalance, where certain odor labels occur far less frequently than others. This imbalance poses a substantial challenge for model training, particularly in accurately predicting rare labels. To address this, we propose an Adaptive Focal Loss combined with regularization strategies to ensure robust generalization when handling imbalanced data. To stabilize training and provide effective probability estimation in the binary classification of odor labels for each molecule, we first employ Binary Cross-Entropy Loss. The formula is as follows:
\begin{equation}
L_{\mathrm{bce}}=-[y\mathrm{log}\left(p_{t}\right)+(1-y)\mathrm{log}\left(1-p_{t}\right)]
\end{equation}
\begin{equation}
p_{t}=\sigma(x)^{y}\cdot(1-\sigma(x))^{(1-y)}
\end{equation}
where \(x\) is the model's raw prediction, \(y\) the ground truth, \(\sigma(x)\) the sigmoid function, and \(p_{t}\) the predicted probability.

To enhance performance on rare labels, we integrate Focal Loss\cite{14}, which focuses on hard-to-classify samples by reducing the contribution of easy samples to the loss. The formula is:
\begin{equation}
L_{\mathrm{focal}}=\alpha\cdot(1-p_{t})^{\gamma}\cdot L_{\mathrm{bce}}
\end{equation}
where \(\alpha = 0.5\) addresses class imbalance by reweighting the importance of positive and negative samples, and \(\gamma = 2\) controls the weight decay of easily classified samples.

A dynamic weighting strategy is adopted to balance the contributions of \(L_{bce}\) and \(L_{\mathrm{focal}}\) during different training stages. In the early stages, greater emphasis is placed on \(L_{\mathrm{bce}}\) to ensure convergence and basic classification ability. Later, the focus shifts to \(L_{\mathrm{focal}}\), improving the model's capacity to handle rare labels and difficult samples. The Adaptive Focal Loss is expressed as:
\begin{equation}
L=\alpha_{1}\cdot L_{\mathrm{focal}}+(1-\alpha_{1})\cdot L_{\mathrm{bce}}
\end{equation}
where \(\alpha_{1}\) is a scaling factor for balancing positive and negative samples, dynamically being increased with training epochs.

To prevent overfitting, \(L_2\) regularization\cite{15} is applied. The final loss function combines these components, effectively addressing data imbalance while ensuring model stability and mitigating overfitting:
\begin{equation}
L_{\mathrm{total}}=L+\lambda \cdot L_2
\end{equation}
where \(\lambda=0.00001\) is the regularization coefficient determined through experimentation.

This design(see Fig.1) optimizes the model across feature encoding, graph convolutional processing, global feature integration, and regularization. It enables the efficient extraction of meaningful features from molecular structures, achieving accurate odor label predictions. The proposed framework demonstrates exceptional performance in multi-label classification tasks, particularly in balancing generalization and predictive power when dealing with complex molecular data.

\section{EXPERIMENTS}

\subsection{Dataset}

We utilized the dataset described in \cite{2}, which provides a robust foundation for investigating the relationship between molecular structures and olfactory descriptors. The dataset comprises 5,788 valid SMILES representations paired with 154 distinct odor descriptors, obtained through meticulous preprocessing to ensure data accuracy and integrity. Each molecule is associated with 1 to over 10 odor descriptors, reflecting its unique olfactory profile (see Fig.2). 80\% of the molecules possess 1 to 6 odor descriptors, while only a small fraction contain more than 10 descriptors, as detailed in Fig.3. Records with missing or invalid SMILES strings were excluded to avoid compromised feature extraction and model training. 

\begin{figure*}[htbp]
\centerline{\includegraphics[width=1\textwidth]{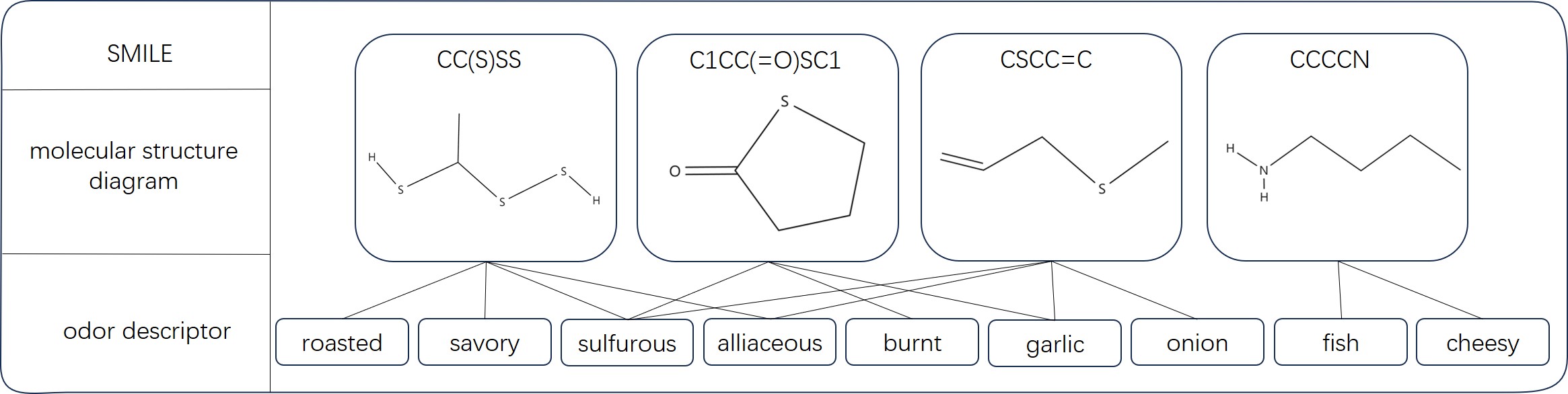}}
\caption{The figure illustrates the SMILES representation of molecular structures in the dataset along with their corresponding odor descriptors, highlighting the relationship between chemical structures and olfactory properties.}
\label{fig}
\end{figure*}
\begin{figure*}[htbp]
\centerline{\includegraphics[width=0.85\textwidth]{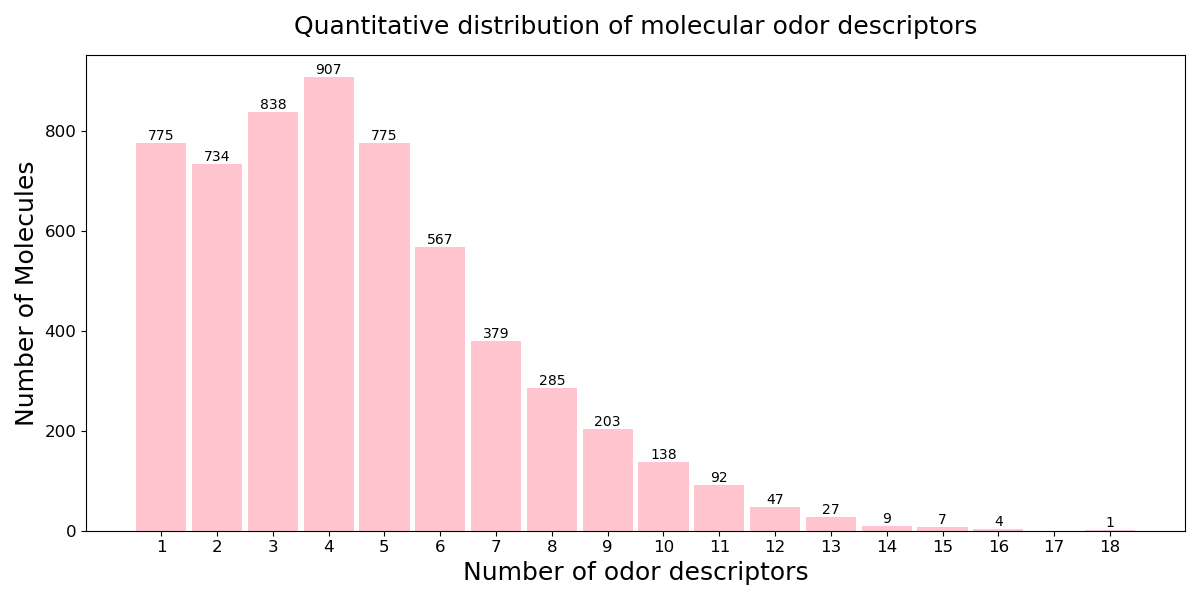}}
\caption{Density distribution of molecular labels in datasets.Eighty percent of the molecules are associated with 1 to 6 odor descriptors, with only a small fraction having more than 10 descriptors.}
\label{fig}
\end{figure*}

\subsection{Method Verification}

\begin{table}[]
\caption{Evaluation results of different models on the dataset, along with ablation experiments for various components.}
\resizebox{0.45\textwidth}{!}{
{\tiny
\begin{tabular}{|c|c|c|}
\hline
Method         & AUROC           & F1 score        \\ \hline
RF+Mordred\cite{x1}(23)     & 0.8382          & 0.3464          \\ \hline
BR+Mordred\cite{3}(22)     & 0.8230          & 0.3555          \\ \hline
CC+Mordred\cite{x8}(24)     & 0.8218          & 0.3513          \\ \hline
KNN\cite{x3}(23)            & 0.7383          & 0.3646          \\ \hline
RNN\cite{x11}(22)            & 0.8321         & 0.3714          \\ \hline
BR+CNN\cite{x10}(24)            & 0.8324          & 0.3758          \\ \hline
GNN\cite{graph}(23)            & 0.9253          & 0.3903          \\ \hline
HAGCN+BCE    & 0.9281          & 0.4475          \\ \hline
HAGCN+AdF     & 0.9292          & 0.4551          \\ \hline
HAGCN+AtA+AdF & \textbf{0.9294} & \textbf{0.4632} \\ \hline
\end{tabular}}}

\vspace{0.5em}
\textit{Note: AtA denotes the Attention Aggregation module, BCE represents Binary Cross-Entropy Loss, and AdF refers to Adaptive Focal Loss. Bold values indicate the best results.}
\end{table}

To validate the effectiveness of the proposed model in the QSOR task, we compared its performance against established methods, including Random Forest (RF) models\cite{x1} based on Mordred descriptors\cite{17}, Binary Relevance (BR)\cite{3}, and Classifier Chains (CC)\cite{x8}. Additionally, we included comparison with Recurrent Neural Networks (RNN)\cite{x11}, K-Nearest Neighbors (KNN)\cite{x3}, BR-based Convolutional Neural Networks (BR-CNN)\cite{x10}, and Graph Neural Network (GNN)\cite{graph}. An 80\%/20\% train-test split was employed to evaluate the model's performance. The primary evaluation metrics were the F1 score and the average Area Under the Receiver Operating Characteristic Curve (AUROC) across odor descriptors. As summarized in TABLE I, the proposed model demonstrates significant performance improvements over traditional Mordred-based approaches by leveraging a multi-level feature extraction strategy that captures comprehensive molecular structure information.  Additionally, our proposed method employs multi-level graph convolutions and attention mechanisms to effectively extract both local and global structural information from molecular graphs. By adaptively weighting adjacent nodes based on their significance, the model captures intricate relationships within molecular structures, resulting in superior predictive accuracy for complex odor profiles.

\subsection{Ablation Study}

We conducted ablation studies on the primary contributions of our method to evaluate the effectiveness of each component.
\begin{table}[]
\caption{In the ablation experiment of different features in molecular feature extraction}
\resizebox{0.50\textwidth}{!}{
{\Huge
\begin{tabular}{|lll|ll|}
\hline
\multicolumn{3}{|c|}{Components}                                                                            & \multicolumn{2}{c|}{Evaluation Metrics}                \\ \hline
\multicolumn{1}{|l|}{Atomic Features} & \multicolumn{1}{l|}{Edge Features} & Fp Features & \multicolumn{1}{l|}{AUROC}           & F1 score        \\ \hline
\multicolumn{1}{|l|}{×}               & \multicolumn{1}{l|}{\checkmark}    & \checkmark                   & \multicolumn{1}{l|}{0.9281}          & 0.4380          \\ \hline
\multicolumn{1}{|l|}{\checkmark}      & \multicolumn{1}{l|}{×}             & \checkmark                   & \multicolumn{1}{l|}{0.9264}          & 0.4529          \\ \hline
\multicolumn{1}{|l|}{\checkmark}      & \multicolumn{1}{l|}{\checkmark}    & ×                              & \multicolumn{1}{l|}{\textbf{0.9356}} & 0.4471          \\ \hline
\multicolumn{1}{|l|}{\checkmark}      & \multicolumn{1}{l|}{\checkmark}    & \checkmark                   & \multicolumn{1}{l|}{0.9294}          & \textbf{0.4632} \\ \hline
\end{tabular}}}

\vspace{0.5em}
\textit{Note:Fp Features represents Fingerprint Features, \checkmark indicates that it contains the feature, and × indicates that it does not contain the feature. Bold values indicate the best results.}
\end{table}

\noindent\textbf{Feature Extraction} To evaluate the effectiveness of multi-level molecular feature extraction, we conducted three ablation experiments by excluding specific feature types: 1) atomic features, 2) edge features, and 3) molecular fingerprint features. As shown in TABLE II, the removal of atomic features caused a significant performance drop, underscoring their importance in predicting complex odor molecule structures, while excluding edge features impaired the model’s ability to represent bond-level interactions. Interestingly, excluding fingerprint features resulted in a higher AUROC but the lowest F1 score, which is likely due to class imbalance and the distinct sensitivity of evaluation metrics to positive sample scarcity in odor prediction tasks. Overall, the multi-level molecular feature extraction strategy effectively integrates diverse molecular information, providing a comprehensive and precise structural representation that improves performance in odor prediction. These results highlight the robustness of our method, which outperforms traditional Random Forest models using Mordred descriptors, even with a reduced feature set.

\noindent\textbf{Network Architecture} To evaluate the effectiveness of the Hierarchical Attention Graph Convolutional Network (HAGCN) architecture, we conducted ablation experiments by systematically removing or replacing key components and analyzing their impact. As shown in TABLE I, first, removing the attention pooling and global feature integration mechanisms led to reduced performance, highlighting the necessity of combining global molecular properties with local graph structural information to enhance classification accuracy. The attention pooling mechanism further amplifies critical node contributions, facilitating more effective feature aggregation. Additionally, the proposed Adaptive Focal Loss addresses class imbalance by dynamically adjusting weights with \(\alpha\) and \(\gamma\), reducing majority-class dominance and enhancing the model's sensitivity to underrepresented odor labels. This dynamic adaptation during training tailors the model to varying sample difficulties, improving its robustness. Collectively, these experiments validate that the proposed architecture, with its attention mechanisms, global-local integration, and Adaptive Focal Loss, significantly advances molecular odor prediction in complex multi-label tasks.

\section{CONCLUSION}

In this paper, we firstly propose a graph attention convolutional network architecture that incorporates a multi-level molecular feature extraction strategy, effectively capturing complex molecular structures and their relationships with odors while addressing the limitations of traditional methods. Moreover, we design the network with multiple graph convolutional layers, enhanced by global feature integration and attention pooling mechanisms, enabling efficient molecular graph representation and significantly improving classification performance across diverse odor descriptors. Finally, we introduce the Adaptive Focal Loss function to address label imbalance in multi-label classification tasks, dynamically adjusting weight distributions to enhance predictive accuracy for underrepresented odor classes. A series of ablation experiments confirm the effectiveness of each component in the framework. Thus, as a whole, this study provides a robust and efficient method for molecular odor prediction while demonstrating the strong potential of graph neural networks in modeling intricate molecular-odor relationships.

\bibliographystyle{IEEEbib}
\bibliography{icme}

\end{document}